# Employee Turnover Prediction: A Cross-component Attention Transformer with Consideration of Competitor Influence and Contagious Effect

Hao Liu, Deakin University, hao.liu@deakin.edu.au

Yong Ge, The University of Arizona, yongge@arizona.edu

## Abstract

Employee turnover refers to an individual's termination of employment from the current organization. It is one of the most persistent challenges for firms, especially those ones in Information Technology (IT) industry that confront high turnover rates. Effective prediction of potential employee turnovers benefits multiple stakeholders such as firms and online recruiters. Prior studies have focused on either the turnover prediction within a single firm or the aggregated employee movement among firms. How to predict the individual employees' turnovers among multiple firms has gained little attention in literature, and thus remains a great research challenge. In this study, we propose a novel deep learning approach based on job embeddedness theory to predict the turnovers of individual employees across different firms. Through extensive experimental evaluations using a real-world dataset, our developed method demonstrates superior performance over several state-of-the-art benchmark methods. Additionally, we estimate the cost saving for recruiters by using our turnover prediction solution and interpret the attributions of various driving factors to employee's turnover to showcase its practical business value.

# 1. INTRODUCTION

Employee turnover refers to an individual's termination of employment from their current organization (Joseph et al. 2015, Muchinsky and Morrow 1980, Price 2001). It is one of the most persistent challenges faced by companies, particularly those in Information Technology (IT) industry that confront high turnover rates (Moquin et al. 2019). For instance, the overall turnover rate for all industries in US has steadily increased from 23.7% in 2015 to 27.9% in 2019 (Bureau of Labor Statistics 2020); Moreover, as reported in the survey (Society for Human Resource Management 2019), 49% of US employees have contemplated resigning from their current employers. Employee turnover is disruptive to the functioning of companies and incurs substantial costs associated with finding and onboarding suitable replacements. These costs encompass expenses related to separation operations, recruitment efforts, lost productivity, the training of new employees (Hausknecht and Holwerda 2013). For example, the Work Institute conservatively estimated the cost to be approximately $15,000 for each employee loss and the total costs associated with voluntary turnover have nearly doubled from $331 billion to $617 billion over the past decade (Mahan et al. 2019).

To alleviate this challenge, it is crucial to develop predictive analytics solutions to analyze and forecast employee turnover, which can benefit multiple stakeholders. First, successfully identifying employees with potential turnover intentions and analyzing possible underlying factors enables employers to address their needs and concerns more effectively to retain them (Mahan et al. 2019, Society for Human Resource Management 2019). Such early intervention may lead to substantial cost savings. Second, an accurate prediction of future employee turnover can help recruiters seek potential candidates for vacant job positions. Today's recruiters primarily search for candidates based on the match between job requirements and candidate qualifications (SocialTalent 2017). But some candidates with good match may have promising careers at their current companies and lack the intention to switch jobs. The prediction of future employee turnover enables recruiters to focus on candidates who demonstrate a good fit for the vacant position and have a high likelihood of changing jobs, thereby saving significant time and resources by minimizing efforts spent on individuals with low chance of switching jobs.

Given the substantial benefits of turnover predictive analytics for both employers and recruiters, prior studies have been conducted in both Information Systems (IS) and other related literature. First, some studies used survey-based approaches to analyze employee turnover and revealed interesting patterns of turnover along different dimensions such as pay gap and gender (Joseph et al. 2015), industry sectors (Joseph et al. 2012), job satisfaction and desire to move (Joseph et al. 2007), and perceived alternatives (Steel 2002). However, it is challenging to apply such survey-based approaches to massive employees across many firms to predict their turnover behaviors. With the development and adoption of various talent management IT systems, including internal HR management systems (e.g., JazzHR) used by firms and publicly available job search platforms (e.g., LinkedIn and Indeed), massive talent data, such as internal employee performance records and public profession profiles, have been accumulated. The second group of studies have utilized such data to analyze and predict talent turnover and movement (Li et al. 2017, Liu et al. 2020). On the one hand, some prior studies have investigated turnover prediction within a single firm by using internal enterprise data (Li et al. 2017, Teng et al. 2019). Given the heterogeneity among firms and the lack or restricted use (e.g., due to data protection regulations such as GDPR) of internal data, these solutions cannot be readily scaled for the turnover prediction of employees across multiple firms. On the other hand, researchers have also studied the talent flow among firms (Hausknecht & Holwerda, 2013; Y. Liu et al., 2020; Wu et al., 2018; Xu et al., 2019; Teng et al., 2021) by using public profession profiles. In these studies, a talent flow network is constructed by aggregating the employee movement among firms, where nodes and links represent firms and aggregated number of transitions, respectively. Because these studies are based on the aggregated talent flow among firms, they have not addressed the movement of individual employees. Besides, while competitor influence has been proved to be an important factor affecting employee turnover (Liu et al. 2020, Oehlhorn et al. 2019), it has been neglected in existing turnover prediction approaches.

To fill the aforementioned research gaps, this paper focuses on predicting individual employees' turnover across different companies by utilizing public career profiles. Specifically, we formulate our research problem as follows: *given a set of career trajectories (i.e., sequences of job positions in employees' working histories) as well as associated employees' and companies' features observed by a specific time, we aim to* *predict the turnover likelihood of the employees in a subsequent period, such as six months*. Tackling this research problem presents several methodological challenges. First, employee turnover is influenced by different types of factors (e.g., employee and company factors). It is unknown and challenging to identify and operationalize these factors from massive semi-structural data. Second, competitor influence plays a significant role in driving employees' turnover decisions (Y. Liu et al., 2020), i.e., employees in one firm may be inclined to leave due to the attraction of other firms. Incorporating such competitor influence into the turnover prediction model poses a great challenge. Third, employees' turnover intentions may be influenced by departed colleagues, i.e., employees may seek other opportunities because their previous colleagues with similar background find better jobs. Integrating such contagious effects into the prediction model is challenging. Fourth, as career trajectories are essentially temporal and sequential data, there is intrinsic dependency among job positions within individual career trajectories. It is critical yet challenging to model the dependency for predicting future turnovers.

To address these challenges, this paper develops a novel Cross-component Attention Transformer with Consideration of Competitor Influence and Contagious Effect (CATCICE) to predict the turnovers of individual employees across different companies, in contrast to extant methods that predict either the turnover within a single organization or the aggregated employee movements among firms, but cannot produce predictions for individual employees across many companies. To achieve this, we follow the design science paradigm (Hevner et al. 2004) to guide the development of the IT artifact. Specifically, we apply Job Embeddedness theory as the kernel theory to identify meta-requirements for the design process and propose meta-designs to satisfy the meta-requirements, and we instantiate an employee turnover prediction system based on the meta-designs.

Our study contributes to the literature in threefold. *First*, we propose a theory-driven design framework for predicting individual employee turnover across multiple firms. In this framework, we identify and operationalize three groups of driving factors rooted in theoretical foundations, and we construct a company graph to assess competitor influence and a title graph to measure contagious influence in employee turnovers. *Second*, we design a novel cross-component transformer model to not only model time-dependencies within career trajectory data but also integrate contagious effect and competitor influence in employee turnovers. The designed model comprises deeply customized cross-component attention and multi-head self-attention neural networks, *which is substantially different from existing transformer models and cross-attention mechanisms* (Vaswani et al., 2017; R. J. Chen et al., 2021; Lu et al., 2019). *Third*, our research underscores the crucial role of kernel theories in design science research, specifically within the domain of talent predictive analytics. Our study employs job embeddedness theory (Batt and Colvin 2011, Rubenstein et al. 2019, William Lee et al. 2014) to guide the development of our predictive method, thus advocating for the use of theoretical frameworks to inform the design of IT artifacts in a rigorous manner (Gregor and Hevner 2013, Li et al. 2020). We conduct extensive evaluations utilizing a real-world dataset and demonstrate superior performance of our method over several state-of-the-art benchmark methods. We also estimate substantial cost savings for recruiters utilizing our turnover prediction solution, thereby validating its practical business value. Furthermore, we illustrate the attribution of the input factors to provide interpretability and support HR managers and recruiters in comprehending the underlying rationale behind the predictions of our solution.

## 2. RESEARCH BACKGROUND

### 2.1 Talent Predictive Analytics

Talent analytics, also known as people analytics, employs statistical or machine learning methods to analyze and solve valuable business problems in human resource management (Marler and Boudreau 2017, Nocker and Sena 2019), such as similar job search (Liu and Ge 2023). This field has evolved with the growing importance of talent predictive analytics (Davenport et al. 2010), which focuses on predicting future talent-related events by analyzing internal and/or external career patterns in the data. The events considered in talent predictive analytics include, but are not limited to employee turnover (Somers and Birnbaum 1999, Teng et al. 2019, Wang et al. 2013), predicting the next employer (Meng et al. 2019), wage prognostication (Vafa et al. 2022), promotions/demotions within a firm (Igbaria and Baroudi 1995, Li et al. 2017), and movements among organizations (Joseph et al. 2012, Liu et al. 2020, Xu et al. 2019). Successful prediction of these events could greatly benefit various talent management tasks including talent acquisition, retention, and development (Ekawati 2019, Sivathanu and Pillai 2019).

Prior research has developed two ways for studying talent-related event prediction. The first one is utilizing internal enterprise data to predict employees' job movements and turnovers within a firm ( Li et al. 2017, Teng et al. 2019, Sun et al. 2019, Hang et al. 2022, Ahuja et al. 2007, Ahuja et al. 2007). The second one uses cross-firm data to analyze and predict aggregated employee movements among firms (Garg et al., 2018; Teng et al., 2021; Xu et al., 2019, Liu et al. 2020). A detailed review of representative studies in both groups is provided in Appendix A-1. However, none of these studies have addressed the prediction of individual employee's turnover across different firms, which is the research focus of this paper.

### 2.2 Employee Turnover Prediction

While extant studies have addressed employee turnover problem from both explanatory and predictive perspectives (Ahuja et al. 2007, Armstrong et al. 2015, Ferratt et al. 2005, Joseph et al. 2007, 2012, Naidoo 2016, Teng et al. 2019, Xu et al. 2019), we concentrate on introducing the related works on turnover prediction because the focus of this paper is centered on the prediction of employee turnover. These related works have studied three groups of technical methods for employee turnover prediction. We summarize the key relevant studies in Table 1 where we compare them with our study from multiple important dimensions.

The *first* stream is survival analysis, which refers to a set of statistical methods for analyzing the time duration until an event of interest occurs (Lee and Wang 2003). Several representative survival analysis models have been applied to turnover prediction (Shen et al., 2010; Chang et al., 2008, Gu et al. 2020, Li et al. 2017). The *second* stream involves traditional classification methods, which treat the turnover prediction as a binary classification problem (Zhao et al. 2019). Traditional classification techniques, such as Logistic Regression and Gradient Boosting Tree (Zhao et al. 2019), have been applied to predict employee turnover. The *third* stream involves modern deep learning methods, which has developed advanced deep learning models to tackle various turnover-related prediction tasks (Teng et al. 2021, Cai et al. 2020, Sun et al. 2019, Meng et al. 2019). Due to space limit, a detailed review of representative studies in the three streams is provided in Appendix A-2.

Compared with these existing works, the novelty of our developed turnover prediction model is *twofold*: *first*, we consider competitor influence and contagious effects and integrate both into our prediction model, which significantly improves the employee turnover prediction accuracy as demonstrated in our evaluation results (see Section 5.4); *second*, we propose a kernel theory-driven design framework and develop a novel cross-component attention transformer model for predicting employee turnover, and our method yields superior performance over the state-of-the-art approaches.

| | Objectives | Methodologies | Single/multiple firms | Time dynamics | Competitor influence | Contagious effect |
|---|---|---|---|---|---|---|
| **Ours** | Predict individual turnover | Cross-component Attention Transformer | Multiple | Yes | Yes | Yes |
| Meng et al. (2019) | Predict the next company & duration | HCPNN | Multiple | Yes | No | No |
| Teng et al. (2021), Xu et al.(2019) | Predict aggregated employee movements | TINN, DSP | Multiple | Yes | No | No |
| Teng et al. (2019) | Predict individual turnover | CEHNN | Single | Yes | No | Yes |
| H. Li et al. (2017) | Predict individual turnover | MTLR | Single | Yes | No | No |
| Zhao et al. (2019) | Predict individual turnover | Decision Tree, Logistic Regression, Random Forest, Support Vector Machines, Gradient Boosting Tree | Single | No | No | No |
| Chang et al. (2008) | Predict individual turnover | COX | Single | Yes | No | No |
| Shen et al. (2010) | Predict individual turnover | AFT | Multiple | Yes | No | No |
| Cai et al. (2020) | Predict individual turnover | DBGE | Multiple | Yes | No | No |
| Hang et al. (2022) | Predict individual turnover | MAHGNN | Single | Yes | No | No |
| Sun et al. (2019) | Predict individual turnover | POFNN | Single | Yes | No | No |

Table 1: Comparison between our study and key relevant studies.

### 2.3 Transformers for Sequential Data

Transformers have achieved state-of-the-art performance for modeling the dynamics of sequential data, particularly text data (Vaswani et al. 2017). They have been utilized in resolving various real-world problems, including customer dialogue learning (Chen et al. 2023) and question answering (Devlin et al. 2019). The core technology of transformer, the self-attention mechanism, is a cutting-edge method to model intra-dependencies within a single sequence (Vaswani et al. 2017). To facilitate the modeling of inter-correlations between two distinct modalities (e.g., text and image or video), the cross-attention mechanism was developed to encompass interactions between two separate sequences (Chen et al. 2021, Lu et al. 2019, Tsai et al. 2019). While our proposed CATCICE model is built upon the core idea of attention mechanism, we have made substantial extensions with novel neural network designs in our CATCICE model. In Section 3.3.5, we provide a summary of the novel extensions by comparing our method with prior work.

## 3. KERNEL THEORY-BASED DESIGN

Our design science procedure comprises four components: kernel theories, meta-requirements, meta-designs, and testable hypotheses (Walls et al. 1992). The kernel theories, derived from the natural and social sciences, govern the meta-requirements. The meta-designs construct the design of information technology (IT) artifacts to satisfy the meta-requirements. Testable hypotheses are employed to evaluate whether the designed artifacts meet the meta-requirements. Our design framework is presented in Table 2 and its details are discussed below.

| Kernel theories | Job embeddedness theory of turnover, which posits that employees remain with their organizations because they are embedded when they have strong links with other people and institutions, perceive a good fit with their jobs and communities, and would incur substantial sacrifices if they were to resign from their jobs, thereby preventing them from leaving their organizations. |
|---|---|
| Meta-requirements | 1. The model should integrate employee factors, company factors, and employee-company interaction factors to predict turnover. It must identify and operationalize these factors to accurately reflect the dimension of "fit" to the company and the dimension of "sacrifice".<br>2. The model must consider the impact of competitor influence on employee turnover decisions. This involves identifying competitors and quantifying their influence, reflecting the dimension of "fit" to the community.<br>3. The model needs to quantify the contagious effect of departed colleagues on employees for accurate turnover prediction. This reflects the dimension of "links".<br>4. The model should account for the dynamic nature of job embeddedness, capturing temporal dependencies among employee factors, company factors, employee-company interaction factors, competitor influence, and contagious effect. |
| Meta-designs | 1. Natural Language Processing (NLP) techniques are used to extract employee factors, company factors, and employee-company interaction factors.<br>2. A competitor graph is constructed to identify competitors, and the High-Order Proximity preserved Embedding (HOPE) method is used to measure competitor influence.<br>3. A title graph is constructed to identify peers with similar functions, and the HOPE method is used to measure the contagious effect.<br>4. A cross-component attention transformer is developed to capture the time dependencies for turnover prediction. |
| Testable hypotheses | Hypothesis 1: Integrating employee factors, company factors, and employee-company interaction factors can improve turnover prediction. |

| | Hypothesis 2: Incorporating competitor influence can improve the accuracy of turnover prediction.<br>Hypothesis 3: Incorporating contagious effect can improve the accuracy of turnover prediction.<br>Hypothesis 4: The designed cross-component attention transformer model can capture the time dependencies to improve turnover prediction. |
|---|---|

Table 2: Overview of our design framework.

### 3.1 Kernel Theory: Job Embeddedness Theory

Job embeddedness theory is one of the most prominent theories for employee turnover analysis (Batt and Colvin 2011, Rubenstein et al. 2019, William Lee et al. 2014), which suggests that an employee's decision to stay with or leave a company is influenced by their degree of embeddedness within the job and company. This theory is grounded in three key dimensions: "links", "fit", and "sacrifice". We posit that job embeddedness theory is an appropriate theoretical foundation to guide the design of a turnover prediction framework.

First, the "fit" in the job embeddedness theory refers to an employee's "perceived compatibility or comfort" with a company and its surrounding communities (William Lee et al. 2014). According to the theory, an employee's attributes (e.g., personal values, skills, career goals, abilities) should align with the demands and features of the company (e.g., culture, values, opportunities, financial status). A better fit indicates a higher likelihood that an employee will feel tied to a company, further contributing to a lower turnover intention (Mitchell et al. 2001). Additionally, the "fit" with the community and surrounding environment is also considered by employees when deciding to stay with or leave their current employers. Although the theory does not explicitly measure the influence of the match between employees and external competing companies on employees' turnover behaviors, as online recruitment become more mature and effective, the external opportunities presented by competing companies have emerged as an important consideration for turnover decisions (Liu et al. 2020, Wright et al. 1994). The competition for human resources among companies is significant because human capital is a key factor for a firm to sustain a competitive advantage (Liu et al. 2020, Wright et al. 1994). Therefore, competitors can influence an employee's turnover intention by presenting themselves as a better fit for potential recruits.

Second, the "sacrifice" dimension in the job embeddedness theory refers to the "perceived loss or sacrifice" associated with leaving a company. The loss could involve giving up colleagues, impactful projects, or stock options, etc. The more an employee would sacrifice by quitting a job, the more difficult it becomes for them to leave the company (Mitchell et al. 2001, Shaw et al. 1998). Opportunities for job stability and advancement, although less visible, are still important potential sacrifices to consider when leaving a company. The more deeply an employee engages with the company (e.g., higher job seniority and longer tenure), the more opportunities for stability and advancement they have, further lowering their turnover intention. For example, marketing managers with more seniority tend to have lower intentions to leave than those with less seniority (DeConinck and Bachmann 1994). And the strength of an employee's job involvement in the company could reflect their perceived sacrifice associated with leaving (Mitchell et al. 2001, Ramlall 2004, Shaw et al. 1998).

Third, "links" dimension in the job embeddedness theory refers to the "discernable connections" an employee has with other people and groups within the company. An employee is more bound to the job and the organization when they have a higher number of links (Mitchell et al. 2001). In addition to the number of links, subsequent studies also investigated the contagious effect of colleagues via social connections (Felps et al. 2009). When frequent turnovers and discussions about turnover occur in a workplace, this turnover contagious effect is likely to prime other employees, even those who are deeply embedded, to consider quitting (Felps et al. 2009). The turnover contagious effect may weaken their confidence about their jobs and the company since many colleagues left the company, and simultaneously strengthen their self-confidence to seek comparable or better job opportunities elsewhere, as previous colleagues switched jobs successfully.

Forth, the dynamic nature of the job embeddedness influences employees' turnover decisions (Holtom and Darabi 2018, Thakur and Bhatnagar 2017). Subsequent studies posits that the assessment of job embeddedness is inherently dynamic (William Lee et al. 2014). First, an employee's perceived "fit" with their job role and the company evolves gradually due to personal growth, changes in job responsibility, growing desires, company culture shifts, etc. A better fit with the company can lead to greater job satisfaction and embeddedness, decreasing the turnover likelihood. Conversely, the likelihood of leaving increases if the "fit" deteriorates over time. Additionally, an employee's "fit" with competitors changes due to personal and external factors. Second, the number and strength of "links" an employee builds within the company tend to increase over time. And the contagious effect changes over time as the number of departed employees fluctuates (Felps et al. 2009). Third, the perceived cost or sacrifice of leaving a company typically increases over time, including loss of career advancement opportunities and benefits. For example, as an employee's service time goes longer, they become more "stuck" to the company, and their turnover intention decreases over time. As perceived sacrifices grow, employees are more likely to stay to avoid these losses.

## 3.2 Meta-requirements

Guided by the job embeddedness theory of turnover and subsequent studies, there are four essential requirements for developing a comprehensive model for predicting employee turnover. First, it demands efforts to incorporate employee factors, company factors, and employee-company interaction factors for turnover prediction. In accordance with the dimension of "fit" to company, it is suggested the alignment between the employee factors and the company factors significantly influences turnover intentions. Consequently, it is imperative to define and operationalize these employee and company factors. Furthermore, the "sacrifice" dimension suggests that a stronger job involvement within an organization indicates a higher perceived cost of leaving, thereby reducing turnover intentions. Therefore, the employee-company interaction factors (e.g., job seniority and length of service) should be integrated into the turnover prediction model.

Second, the dimension of "fit" to community implies that the influence of competing companies is pivotal in employees' turnover decisions. Hence, it is essential to incorporate competitor influence into the turnover prediction model. To achieve this, the designed model should initially identify the competitors of an employee's current company in terms of human capital competition (Liu et al. 2020) and subsequently measure the competitive influence exerted by alternative companies.

Third, the model must account for the contagious effect on focal employees originating from colleagues who have departed from the company. The dimension of "links" to colleagues suggests that turnover contagion may weaken an employee's confidence in their job and company, thereby promoting their inclination to pursue better opportunities elsewhere. The contagious effects on a focal employee vary contingent upon the strength of the connection between the departed colleague and the focal employee.

Therefore, to quantify the contagious effect, the designed model should measure the strength of the "connection" between a focal employee and a departed employee, and aggregate the cumulative contagious effects emanating from departed colleagues.

Fourth, considering the dynamic nature of job embeddedness, it is imperative to account for temporal dependencies of employee factors, company factors, employee-company factors, competitor influence, and contagious effects in the turnover prediction model. This is because changes in job embeddedness are significant predictors of employee turnover behavior.

### 3.3 Meta-designs

In this section, we introduce our meta-designs for an effective turnover prediction system, which addresses the meta-requirements discussed in the preceding section. An overview of our design is illustrated in Figure 1. We collected public employee profiles and company financial statements as input data. Let $U$ denote a set of $N$ employees and let $C$ denote a set of $M$ companies that appear in employees' employment records, with an observation period spanning $T$ months. The employee turnover prediction problem studied in this paper is formulated as follows: given a set of employees' career trajectories observed within a specific time period $\tau_o$, our objective is to predict the employees' turnover likelihood during a subsequent time period $\tau_p$ (e.g., the ensuing 6 months). We commence by describing the factor identification and operationalization for employee factors, company factors, and employee-company factor. Subsequently, we introduce the integration of competitor influence and contagious effect. Finally, we present a novel cross-component attention transformer to model the sequential data for solving the turnover prediction problem. We discuss our meta-designs from four aspects in the following.

### 3.3.1 Meta-design I: Identify and operationalize employee factors, company factors, and employee-company factors for employee turnover prediction.

Previous studies on employee turnover have revealed a variety of factors that engender turnover intentions and behaviors. We identified and operationalized multiple factors within each category of employee factors, company factors, and employee-company factors (Mitchel 1981) to constitute a concise set of representative features. These features will be supplied as inputs for our developed deep learning model. Let $\boldsymbol{x}_{ui}^E, \boldsymbol{x}_{ui}^C, \boldsymbol{x}_{ui}^{EC}$ denote the operationalized features of employee factors, company factors, and employee-company factors of the employee $u$ at the $i$-th month, respectively. In the remainder of this section, we introduce the three categories of factors and their operationalized features. Furthermore, a summary of these factors is presented in Table 3 and implementation details are provided in Section 4.

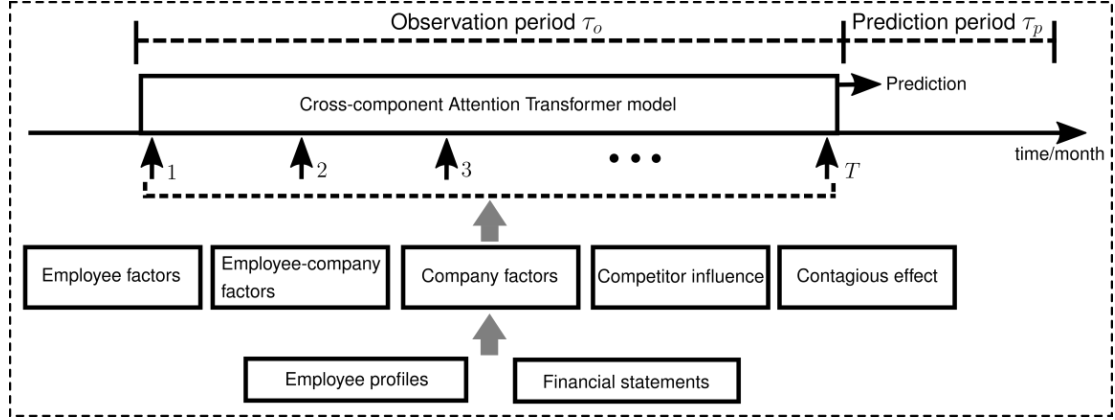

Figure 1: Design overview of cross-component attention transformer model.

**Employee Factors**

Employee factors refer to the individual characteristics and preferences such as gender, education level, experience, needs and ambition (Joseph et al. 2015, Ng et al. 2007, Ramlall 2004). These individual factors influence employees' preferences for and behaviors associated with job turnover (Ng et al. 2007).

The career length, which represents the duration an employee has been employed, may influence turnover behavior in multiple ways. First, human capital theory suggests that as the career length increases, employees acquire more experience and knowledge, rendering them more marketable and capable of turnover (Joseph et al. 2007). Second, the career mobility patterns vary temporally. On one hand, employees would expect a rapid job progression at early stages, and subsequently, their careers stabilize gradually with higher wages (Manzoni et al. 2014). On the other hand, career stage and development theories suggest that as employees age, they exhibit lower turnover intentions since they are more satisfied with their jobs (Joseph et al. 2007). To this end, we define the *Number of total working months* as one feature representing the career length for turnover prediction.

*Gender* is another salient factor influencing turnover behaviors as female and male employees exhibit different job movement patterns (Joseph et al. 2015). Female employees often reduce working hours and even suspend their careers due to family reasons such as childbearing (Sorensen et al. 1999). Meanwhile, previous studies also found that female employees are perceived to have fewer promotion opportunities than males (Igbaria and Baroudi 1995, Joseph et al. 2007). While the discrepancy effect is diminishing over time, gender remains an important feature that needs to be considered for predicting turnover events.

Education background plays as an significant role in influencing turnover (Iverson and Pullman 2000, Joseph et al. 2015, Kronenberg and Carree 2012, Sicherman and Galor 1990). Employees with higher educational levels usually expect more opportunities for job progression (Sicherman and Galor 1990). According to human capital theory, when employees with higher education are dissatisfied with their jobs, they will turn over to other companies quickly because they are adept at collecting information on alternative jobs and more proficient in utilizing their knowledge in different environments (Kronenberg and Carree 2012, Ramlall 2004). Concurrently, companies tend to retain employees with higher education because they usually possess greater human capital (Iverson and Pullman 2000). Therefore, we consider *The highest education degree* as one feature toward predicting employees' turnovers.

Additionally, prior studies have revealed that the frequency of changing jobs is a significant predictive variable for turnover (Joseph et al. 2007, Ng et al. 2007, Vianen et al. 2003). For instance, employees who frequently change their jobs often have needs to pursue new challenges or rapid career promotions because they seek to transition to other companies when their needs are unmet. Therefore, we define multiple features to measure employees' preferences for such needs, including *Number of turnovers in the past*, *Average working months in previous companies* and *Number of position changes in the past*, and consider them for turnover prediction.

**Company Factors**

Company factors measure the characteristics of a company that affect employees' turnover. First, the company size influences the tendency of employees' turnover because larger companies usually provide better benefits and career opportunities, and have lower turnover rate (Kronenberg and Carree 2012). Thus, we define *Number of total employees* reflecting the size of a company and consider it for turnover prediction. Second, company expansion implies the addition of more positions, which could attract workers from other companies (Ng et al. 2007). To this end, we define *Number of new hired employees* and *Number of separations* to measure size trend (expansion or downsizing) of a company and include both measurements in our turnover prediction model. Third, financial performance affects companies' investment in employee relation policies and working environment, which consequently influences employees' satisfaction and turnover intention (Ng et al. 2007, Waddock and Graves 1997); poor financial performance is one of the most salient reasons for employment downsizing such as layoffs, deemed as an effective way to reduce costs and improve financial situations (Cascio et al. 1997, Datta et al. 2010). The financial performance of a company could be measured by the following variables derived from financial statements: *Revenue*, *Cost of revenue*, *Net income*, *Earnings per share (EPS)*, *Cash and cash equivalents*, *Total assets*, *Total debt*, *Total liabilities*, *Stock-based compensation*, *Operating cash flow*, *Investing Cash flow*, and *Financing cash flow* (Revsine et al. 2021), all of which are considered for turnover prediction in this paper.

**Employee-company Factors**

Employee-company factors reflect the interaction between employees and companies (Muchinsky and Morrow 1980) that influence turnover. First, location has been identified as a salient factor influencing employees' turnover. The turnover rate is often substantially higher in large metropolitan areas than in small urban and rural places due to more employment opportunities in these locales (Kronenberg and Carree 2012, McCollum et al. 2018). We thus define the *location* feature (i.e., state, city) for turnover prediction. Second, organizational commitment theory suggests that employees' commitment, which is the involvement strength of an employee in an organization, influences turnover (DeConinck and Bachmann 1994, Ramlall 2004). Organizational commitment tends to increase as an employee's service time and job seniority increase in the organization, thereby lowering their turnover intention (DeConinck and Bachmann 1994, Ramlall 2004); and employees with high seniorities often have to sacrifice more when they leave their current companies (Joseph et al. 2007). Therefore, we define *Number of months working in current company* and *Job seniority* to measure employees' organizational commitment in our turnover prediction model. Third, *Job title* in current company, which indicates job characteristics, is also related to employees' turnover (Joseph et al. 2015, Ramlall 2004). Thus, we include it as another feature for turnover prediction.

| Category | Definition | References |
|---|---|---|
| Company factors | Number of total employees | Kronenberg & Carree (2012) |
| | Number of new hired employees | Ng et al. (2007) |
| | Number of separations | |
| | Financial features: Revenue, Cost of revenue, Net income, Earnings per share (EPS), Cash and cash equivalents, Total assets, Total debt, Total liabilities, Stock-based compensation, Operating cash flow, Investing cash flow, Financing cash flow | Cascio et al. (1997), Datta et al. (2010), Ng et al (2007), Waddock and Graves (1997) |
| Employee factors | Gender | Igbaria and Baroudi (1995), Joseph et al. (2007, 2015), Sorensen et al. (1999) |
| | The highest education degree | Iverson & Pullman (2000), Kronenberg and Carree (2012), Ramlall (2004), Sicherman and Galor (1990) |
| | Number of total working months | Joseph et al.(2007), Manzoni et al. (2014) |
| | Number of job turnover in the past | Joseph et al. (2007), Ng et al. (2007), Vianen et al. (2003) |
| | Average working months in previous companies | |
| | Number of position changes in the past | |
| Employee-company factors | Location of current job (State, City) | Kronenberg and Carree (2012), McCollum et al. (2018) |
| | Job title | Joseph et al. (2015), Ramlall (2004) |
| | Job title seniority (Owner, Chief X Officer, Vice President, Director, Manager, Senior, Entry-Level) | DeConinck and Bachmann (1994), Joseph et al. (2007), Ramlall (2004) |
| | Number of months working in current company | |

Table 3: A summary of operationalization of employee factors, company factors, and employee-company factors.

### 3.3.2  Meta-design II: Measure and integrate company competitor influence for turnover prediction.

To incorporate the competitor influence in predicting employee turnover, we construct an employee flow graph to identify company competitors, then we aggregate the competitors' factors and integrate them into the predicting model. First, we create an employee flow graph based on all employee career trajectories, where each node represents a company and each edge connecting two nodes represents job movement between these two companies. The weight of each edge quantifies the frequency of the job movements. Let $G_c(V_c, E_c)$ denote the employee flow graph, where $G_c$ is an directed graph, $V_c$ is the set of company nodes and $E_c$ represents the set of weighted edges. Each edge weight is generally treated as the measure of similarity between the two nodes because a larger number of movements between two companies indicate a higher substitutability of the two companies. We employ High-Order Proximity preserved Embedding (HOPE) (Ou et al. 2016) to learn the company embeddings from the employee flow graph. Let $f_{HOPE}(\cdot)$ denote the HOPE operation, and $\{e_1^C,\ e_2^C, \ldots, e_M^C\}$ represent $M$ embeddings of the companies from the employee flow graph $G_c(V_c, E_c)$. Then we have:

$$\{e_1^C, e_2^C, \ldots, e_M^C\} = f_{HOPE}(G_C). \tag{1}$$

Subsequently, we select the most similar competitors from the neighbors of each company based on the cosine similarity of a pair of company embeddings from $\{e_1^C,\ e_2^C, \ldots, e_M^C\}$. The competitor influence is measured as the aggregated company features of similar competitors. Let $x_{ci}^{CI}$ denote the competitor influence features in the $i$-th month for company $c$ and $S^{company}(c)$ represent the set of most similar neighbors of company $c$. Then $x_{ci}^{CI}$ is defined as:

$$x_{ci}^{CI} = \frac{\sum_{o \in S^{company}(c)} x_{oi}^C \cdot similarity(c, o)}{\sum_{o \in S^{company}(c)} similarity(c, o)}, \tag{2}$$

where $x_{oi}^C$ is the company features (introduced in section **Error! Reference source not found.**) of the c competitor $o$ in the $i$-th month and $similarity(c, o)$ measures the cosine similarity between the embeddings of companies $c$ and $o$.

### 3.3.3 Meta-design III: Measure and integrate contagious effect of departed colleagues for turnover prediction.

Employee turnover intentions are influenced by the departure of colleagues (Felps et al. 2009, Hang et al. 2022, Teng et al. 2019). To investigate such contagious effects, we construct a job title similarity graph to facilitate identifying influential "peers" and then aggregate the employee factors of the peers who have recently left the company. It is challenging to observe the relationships between the focal employee and previous colleagues from public career profiles. As a resolution, we identify influential peers from a broader relationship based on the similar job functions since a company typically provides comparable policies and opportunities for roles with similar functions, and competitors similarly target these individuals with analogous external opportunities (Joseph et al. 2015, Ramlall 2004). Thus, employees would pay more attention to peers with similar functions in terms of internal and external opportunities. Therefore, we construct a directed job title graph to find titles with similar functions. We observe that movements along a career trajectory can indicate functional similarity between the predecessor and successor titles. Accordingly, we build a title flow graph based on these title movements, where a node represents a unique title and a weighted edge connects two titles, reflecting the frequency of the transitions. It is noteworthy that title movements can occur both internally (e.g., through internal transfers and promotions) and between different companies. Let $G_T(V_T, E_T)$ represent the directed title graph, where $V_T$ is the set of titles and $E_T$ is the set of edges in $G_T$. We then utilize the HOPE method to derive title embeddings from the title graph. Let $f_{HOPE}(\cdot)$ denote the HOPE operation and $\{e_1^T, \ e_2^T, ..., e_Q^T\}$ represent the embeddings of titles in the title graph $G_T(V_T, E_T)$. Thus, we have:

$$\{e_1^T, e_2^T, ..., e_Q^T\} = f_{HOPE}(G_T). \tag{3}$$

To identify influential peers for each employee, we group the departed colleagues by their quitting month. For each month, we select the departed colleagues whose job titles are most similar to that of the focal employee, based on the cosine similarity of their respective title embeddings. The contagious effect for each month is quantified by aggregating employee factors. Let $\boldsymbol{x}_{ui}^{CE}$ denote the contagious effect features for employee $u$ in $i$-th month and $S_i^{title}(u)$ represent the set of peers with the most similar job titles for employee $u$ in the $i$-th month. Then $\boldsymbol{x}_{ui}^{CE}$ is defined as:

$$\boldsymbol{x}_{ui}^{CE} = \frac{\sum_{p \in S_i^{title}(u)} \boldsymbol{x}_{pi}^E \cdot similarity(u, p)}{\sum_{p \in S_i^{title}(u)} similarity(u, p)}, \tag{4}$$

where $\boldsymbol{x}_{pi}^E$ are the employee factors (introduced in section **Error! Reference source not found.**) of employee $p$ in $i$-th month and $similarity(u, p)$ is the cosine similarity between the title embeddings of employees $u$ and $p$.

### 3.3.4 Meta-design IV: Design a novel deep learning model to capture the dynamics of employee factors, company factors, employee-company factors, competitor influence, and contagious effects for turnover prediction.

Inspired by Transformer model (Vaswani et al. 2017), which was designed for modeling sequential dependency of text data, and cross-attention (Lu et al. 2019), which facilitates the exchange of key-value pairs between two modalities (e.g., text and video), we design the architecture of our CATCICE model that is graphically illustrated in Figure 2. We create features and train the model within the observation period ($T$ months) and predict each employee's turnover probability in the prediction period. The features include company factors, employee factors, employee-company factors, competitor influence features, and contagious effect features, all of which are created from two data sources (i.e., publicly-available employee profiles and company financial statements) as introduced in Section **Error! Reference source not found.**- REF _Ref180104978 \n \h \* MERGEFORMAT 3.3.3.

As shown Figure 2, our CATCICE model consists of five components to model the five inputs: employee factors, company factors, employee-company factors, competitor influence, and contagious effect. Each component is composed of a stack of $L$ identical layers, where each layer features a masked multi-head self-attention mechanism and a cross-component attention, followed by two fully-connected layers forming the feed-forward network. The masked multi-head self-attention and cross-component attention are used to capture the intra-component and inter-component time dependencies, respectively. Let $\boldsymbol{Y}_u^l = \{\boldsymbol{y}_{ui}^l\}_{i=1}^5$ represent the five inputs to the $l$-th layer ($0 \leq l \leq L$) for employee $u$. For $l = 0$, $\boldsymbol{Y}_u^0 = \{\boldsymbol{y}_{ui}^0\}_{i=1}^5 = \{\boldsymbol{x}_u^C, \boldsymbol{x}_u^E, \boldsymbol{x}_u^{EC}, \boldsymbol{x}_u^{CI}, \boldsymbol{x}_u^{CE}\}$. The output of masked multi-head self-attention for layer $l$, $\boldsymbol{Z}_u^l = \{\boldsymbol{z}_{ui}^l\}_{i=1}^5$, is computed as:

$$\boldsymbol{z}_{ui}^l = \text{LayerNorm}\left(\text{MaskedMultiHeadAttention}(\boldsymbol{y}_{ui}^{l-1})\right), \tag{5}$$

where $l \geq 1$, and $\text{LayerNorm}(\cdot)$ is layer normalization operation (Ba et al. 2016). $\text{MaskedMultiHeadAttention}(\cdot)$ is the masked multi-head self-attention mechanism, introduced in the decoder of the transformer model (Vaswani et al. 2017). The outputs $\{\boldsymbol{z}_{uj}^l\}_{j=1}^5$ is then processed through the cross-component attention, yielding $\boldsymbol{h}_{ui}^l$ ($i = 1,2,3,4,5$) as:

$$\boldsymbol{h}_{ui}^l = \text{CrossComponentAttention}\left(\{\boldsymbol{z}_{uj}^l\}_{j=1}^5, i\right), \tag{6}$$

where $\text{CrossComponentAttention}(\cdot, i)$ is the cross-component attention mechanism on the $i$-th component. This is followed by two skip connections (He et al. 2016), further layer normalizations and a feed-forward network.

$$\boldsymbol{g}_{ui}^{l} = \text{LayerNorm}\big(\boldsymbol{y}_{ui}^{l-1} + \boldsymbol{h}_{ui}^{l}\big), \tag{7}$$

$$\boldsymbol{y}_{ui}^{l} = \text{LayerNorm}\left(\boldsymbol{g}_{ui}^{l} + \text{FFN}\big(\boldsymbol{g}_{ui}^{l}\big)\right), \tag{8}$$

where $\text{FFN}(\cdot)$ is a network comprising two fully-connected layers. The outputs of the five components are then fused through a linear combination in the multi-component fusion layer:

$$\boldsymbol{O}_u = \sum_{i=1}^{5} \boldsymbol{W}_i \boldsymbol{y}_{ui}^{L} + \boldsymbol{b}, \tag{9}$$

where $\boldsymbol{O}_u$ is the output of the muti-component fusion layer; $\boldsymbol{W}_i$ denotes the weight matrix for the $i$-th component, and $\boldsymbol{b}$ represents the bias term. Finally, the representation of the last month is transformed into a scalar variable by two fully-connected layers (FC) and then is used to estimate the turnover likelihood $\hat{r}_u$ via logistic sigmoid function $\sigma$ for employee $u$:

$$\hat{r}_u = \sigma(FC(\boldsymbol{O}_{uT})), \tag{10}$$

where $\boldsymbol{O}_{uT}$ is the representation of the last month in $\boldsymbol{O}_u$. The loss function $\mathcal{L}_{\text{u}}$ for employee $u$ is defined as:

$$\mathcal{L}_{\text{u}}(\theta) = -[r_u log(\hat{r}_u) + (1 - r_u) \log(1 - \hat{r}_u)] + \beta ||\theta||^2, \tag{11}$$

where $\theta$ represents all learnable parameters and $r_u$ is the true label for employee $u$. And $\beta$ is the hyper-parameter for the $L_2$ regularization penalty $||\theta||^2$.

We show the detailed operation of the formula $\boldsymbol{h}_{u2}^{l} = \text{CrossComponentAttention}\left(\{\boldsymbol{z}_{uj}^{l}\}_{j=1}^{5}, 2\right)$ for employee factors in Figure 3. The Query matrix $Q_2$ for employee factors is used to interact with the Key matrix $K_i$ ($i = 1,2,3,4,5$) of all five components to calculate the attentions $A_{2,i} = Q_2 \cdot K_i^T / \sqrt{d}$ ($i = 1,2,3,4,5$), with $d$ representing the dimension of $Q_2$. And $A_{2,i}$ are subsequently utilized to multiply the corresponding Value matrices $V_i$ ($i = 1,2,3,4,5$). The concatenated outputs from these operations are then fed into a fully-connected layer to derive the embedding $\boldsymbol{h}_{u2}^{l}$. Similarly, this approach is applied to learn the embeddings $\boldsymbol{h}_{u1}^{l}, \boldsymbol{h}_{u3}^{l}, \boldsymbol{h}_{u4}^{l}, \boldsymbol{h}_{u5}^{l}$ for company factors, employee-company factors, competitor influence, and contagious effect, respectively. Such a novel design captures the important inter-component time dependency for turnover prediction.

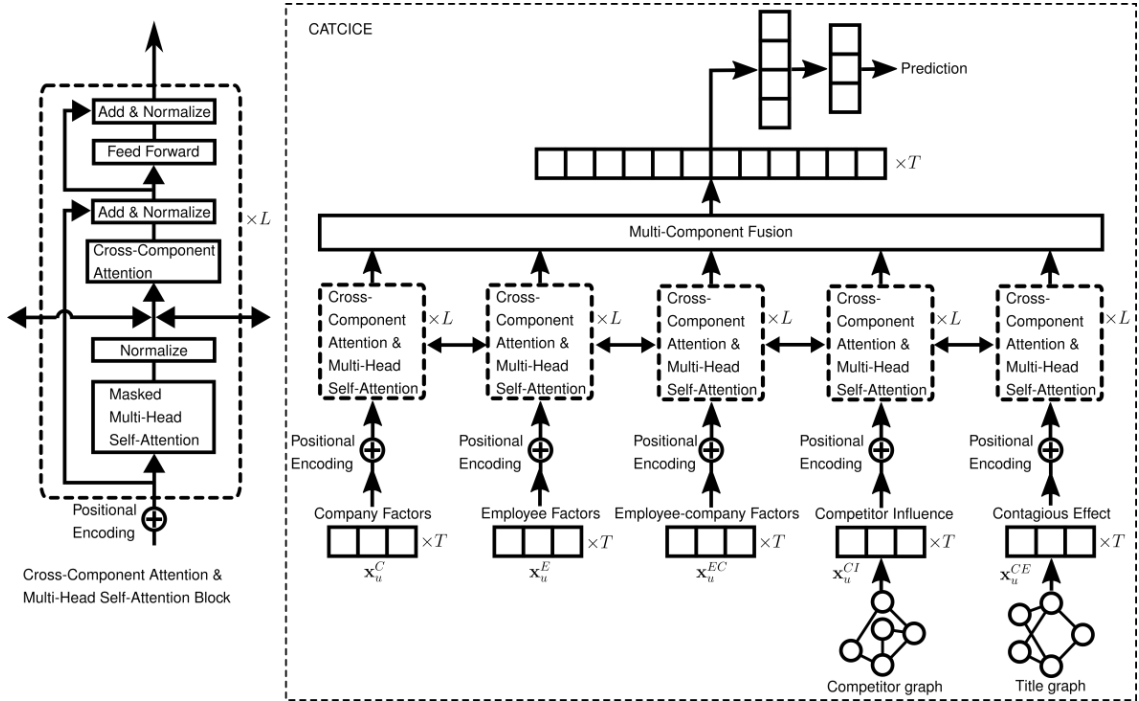

Figure 2: The architecture of our CATCICE model.

### 3.3.5 Remarks on the Novelty of CATCICE Model

Our CATCICE model design fundamentally differs from the well-known Transformer model (Vaswani et al. 2017) because it contains five components and cross-component attention as shown in Figure 2 and 3, which are not considered in the Transformer model. Furthermore, our cross-component attention differs from the cross-attention in previous studies (Chen et al. 2021, Lu et al. 2019, Tsai et al. 2019) *in three aspects*. *First*, previous studies utilized cross-attention between two modalities and switched the Key and Value matrices between them. Instead, we employ cross-attention across five components, and each component absorbs Key and Value of the other four components. We further concatenate and compress them using a fully-connected layer. *Second*, we combine the self-attention with the cross-attentions from the other four components, thus preserving the Value of each component for subsequent layers. *Third*, we project the output of the masked multi-head self-attention into a uniform dimension across five components to prevent any single input from dominating the calculation of cross-component attention due to its large dimensionality. We further add the skip connection from the input of the masked multi-head self-attention to the output of cross-component attention to address degradation issues (He et al. 2016).

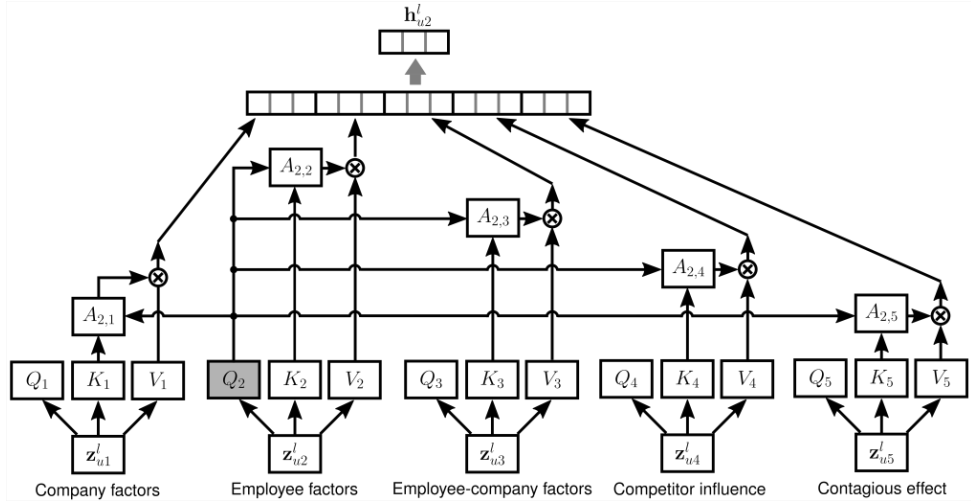

Figure 3: An illustration of cross-component attention for employee factors.

### 3.4 Testable Hypotheses

Testable hypotheses are intended for rigorously assessing whether our meta-designs satisfy the meta-requirements as outlined by Walls et al. (1992). We have formulated four testable hypotheses corresponding to our four meta-designs, which we will examine in Section **Error! Reference source not found.**.

**Hypothesis 1:** The integration of employee factors, company factors, and employee-company interactions enhances the accuracy of predicting employee turnover.

**Hypothesis 2:** The inclusion of competitor influence increases the accuracy of turnover predictions.

**Hypothesis 3:** Accounting for the contagious effects within employee interactions improves the accuracy of turnover predictions.

**Hypothesis 4:** The cross-component attention transformer model represents a superior design for capturing the temporal dynamics of the five components. It is hypothesized to enhance the accuracy of employee turnover prediction compared to existing machine learning approaches.

### 4. DESIGN INSTANTIATION: EMPLOYEE TURNOVER PREDICTION SYSTEM

Here we present the designed artifact: an employee turnover prediction system that includes data preparation, data extraction, machine learning model training, and turnover prediction, as illustrated in Figure 4.

**Data Preparation:** This stage involves the collection of semi-structured employee profiles and financial statements from multiple sources. Employee profiles are sourced from a third-party database, while financial statements are obtained through web crawling.

**Data Extraction:** This step aims to construct features including employee factors, company factors, employee-company factors, competitor influence, and contagious effect from the semi-structured employee profiles and financial statements, as prescribed by the meta-designs. Regarding company features, the historical numbers of overall employees, new hired and separated employees of each company are estimated from professional profiles. Financial features are derived from the financial reports. For employee features, gender is inferred from employees' first names using gender-guess[1] and encoded into a one-hot vector. The highest education degree, which could be PhD, Master, Bachelor, Associate or Other, is also encoded into a one-hot vector. As for employee-company features, we apply binary encoding to represent state and city locations. The representation of each job title is learned by the Sentence-Transformer model (Reimers and Gurevych 2019). Job title seniority is determined using regular expressions based on LinkedIn Seniority Codes and encoded using one-hot encoding. For competitor influence, we select the top ten similar competitors for each focal company to aggregate their features. For contagious effect, we aggregate the features of departed colleagues who held the top ten job titles that are most similar to those of the focal employee. Finally, all features, except for one-hot ones, are normalized using standardization due to the substantial variation in the ranges of their values.

**Model Training:** We train the proposed model on the training dataset and search for the best hyper-parameters on the validation set by following the standard supervised machine learning procedures. Due to space limit, we include the detailed training algorithm in Appendix A-3.

**Turnover Prediction:** We predict the turnover likelihood of each employee using the trained model and visualize the attribution of input features for interpretability.

## 5.  EVALUATION

A core tenet of the computational design science research is the rigorous evaluation of the developed IT artifact against the state-of-the-art benchmark approaches (Hevner et al. 2004). Therefore, we evaluate our developed method by comparing it with several selected benchmark methods. We conduct evaluations and comparisons by using a large-scale real-world data set and multiple evaluation metrics. Furthermore, we conduct a performance attribution analysis to investigate the reliability of our method and provide a simulation study to validate the business value of the developed IT artifact. Lastly, we demonstrate multiple case studies for interpretation analysis of our model.

---

[1] https://test.pypi.org/project/gender-guesser/

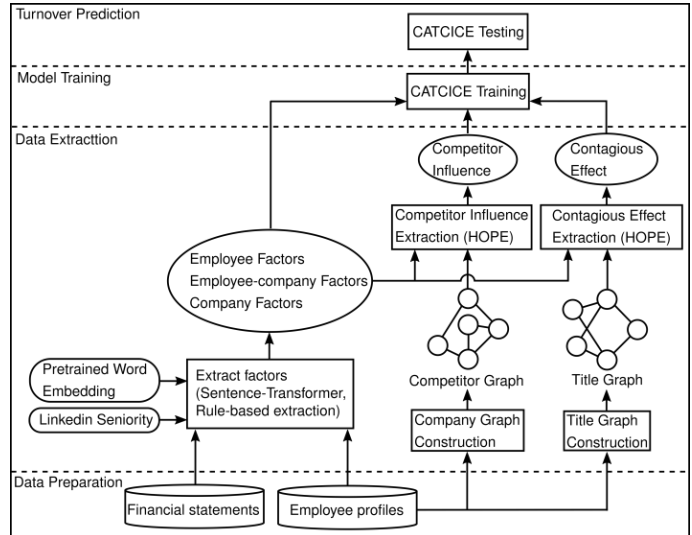

Figure 4: The employee turnover prediction system architecture.

## 5.1 Evaluation Setting

### 5.1.1 Dataset

We utilized a large-scale real-world data set to evaluate the performance of our method. The data set comprises two parts: employees' professional profiles and firms' financial statements. The employees' professional profiles include turnover events, employee and company related features. The firms' financial statements are used to generate financial features about firms. In the following, let us introduce more details about data collection.

The employees' professional profiles were collected in March 2017 from one of the largest professional networking sites in the United States. The employees' professional profile contains full name, industry sector, employment history, and educational background. Employment history is essentially a chronological sequence of job positions. Each job position usually consists of company name, job title, start date, end date, a brief description, and location. A turnover event is identified when one employee transitions from their current company to a different one. This study specifically targeted employees within IT-related sectors due to these sectors exhibiting the highest turnover rates among all industries (Lounsbury et al. 2007, Moquin et al. 2019). To avoid a cold start, we keep employee profiles that contain at least two employment jobs including the current job. Consequently, our dataset comprises 87070 employees, who have worked in 753 companies. We collected these companies' quarterly financial statements from Stockrow[2] and 10-Q forms (quarterly financial statements) from the U.S. Securities and Exchange Commission[3]. In Appendix A-4, we provide some basic statistics about the dataset.

Given that the collected financial statements span from June 2009 and the employee profiles extend until March 2017, we have designated the period from January 2010 to December 2016 for evaluating our developed method. A six-month time window is often needed for companies to retain employees and for recruiters to search for candidates, as suggested by a survey with over 2,000 U.S. professionals in 2018 (Randstad 2018). Consequently, a 6-month duration is proper as the prediction period. Furthermore, we add a second testing period (12 months) to examine our method. A turnover event is characterized by an employee's transition from one employer to another within the designated prediction period. All features in Table 3 are derived from the historical data for each month within the observation period.

We split the data into training, validation, and test sets over time to rigorously evaluate the developed method, as demonstrated in Figure 5. The training set is used to train the learning model; the optimal hyper-parameters are identified using the validation set; and the test sets are used to evaluate the model performance. Specifically, we created two test sets, $Test_6$ and $Test_{12}$, to examine the prediction performance of our model for both 6-month and 12-month intervals. *Since there is no overlap between the prediction periods of training, validation, and test sets, there will be no data leakage issue*. Practically, the turnover prediction model should be built using historical data and the built model is applied to predict future turnovers. This temporal splitting approach aligns closely with real-world practices. The observation period, prediction period, numbers of turnovers, survivals, and overall cases for each of the training, validation, and test sets are shown in Table 4. The statistics of the competitor graph and title graph are presented in Table 5.

---

### 5.1.2 Benchmark Methods

Based on an extensive review of the literature (see Section 2.2), we selected eleven state-of-the-art benchmark methods and evaluated our method against them. These benchmark methods range across survival analysis, traditional classification methods, recurrent neural network, and graph embedding model, summarized in Table 6. These methods were also optimized on the validation set. Due to space limit, we describe each of these benchmark methods in Appendix A-6.

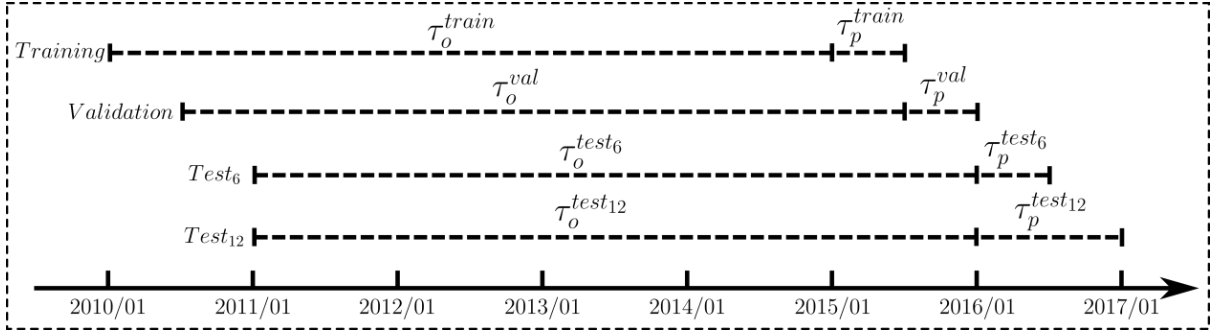

Figure 5: An illustration of training, validation, and test data via the split over time.

| Datasets | Observation period | Prediction period | # of Turnover | # of Survival | # of All Cases |
|---|---|---|---|---|---|
| Training | 2010/01-2014/12 | 2015/01-2015/06 | 6273 | 78857 | 85130 |
| Validation | 2010/07-2015/06 | 2015/07-2015/12 | 6628 | 77740 | 84368 |
| Test$_6$ | 2011/01-2015/12 | 2016/01-2016/06 | 5831 | 77031 | 82862 |
| Test$_{12}$ | 2011/01-2015/12 | 2016/01-2016/12 | 9461 | 73401 | 82862 |

Table 4: Training, validation, and test sets.

| Graphs | # nodes | # edges |
|---|---|---|
| Competitor graph | 753 | 88108 |
| Title graph | 6100 | 19815 |

Table 5: Competitor graph and title graph information.

| Category | Methods | Abbreviation |
|---|---|---|
| Survival Analysis | Cox Proportional-Hazards Model | COX |
| | Log-Logistic Accelerated Failure Time | AFT |
| | Multi-Task Logistic Regression | MTLR |
| Classification | Decision Tree | TREE |
| | Logistic Regression | LR |
| | Support Vector Machines | SVM |
| | Random Forest | RF |
| | Gradient Boost Tree | GBT |
| Recurrent Neural Network | Gated Recurrent Unit | GRU |
| | Hierarchical Career-Path-aware Neural Network | HCPNN |
| Graph Embedding | Dynamic Bipartite Graph Embedding | DBGE |
| Transformer-based model | Cross-component Attention Transformer | **CATCICE** (ours) |

Table 6: Methods compared in the experiments.

### 5.1.3 Evaluation Metrics

The turnover and survival cases are designated as positive and negative instances, respectively. To compare the performance of the competing methods, we adopt three widely-used metrics to evaluate their performance: top-K precision (Precision@k), top-K Normalized Discounted Cumulative Gain (NDCG@k), and Area Under the Receiver Operating Characteristic Curve (AUC). A detailed introduction of the three metrics is provided in Appendix A-5. A summary of the metrics is provided in Table 7. Together, these metrics provide complementary local and global evaluations of prediction performance. *The scale of all metrics ranges from 0 to 1 and a higher value indicates better performance.*

| Metrics | Number Count | Rank Count | Class Discrimination | Top-K List | Whole List |
|---|---|---|---|---|---|
| Precision@k | ✓ | | | ✓ | |
| AUC | | | ✓ | | ✓ |
| NDCG@k | | ✓ | | ✓ | |

Table 7: A summary of evaluation metrics.

## 5.2 Prediction Performance of Different Methods

The evaluation results, which encompass seven specific metrics (i.e., AUC, Precision@30, Precision@50, Precision@100, NDCG@30, NDCG@50, and NDCG@100), are presented in Table 8 and Table 9 for 6-month and 12-month prediction periods, respectively. After tuning on the validation set, the optimal architecture of our model consists of two layers ($L = 2$) and four heads for the multi-head self-attention. We summarize the following observations from the evaluation results.

First, our proposed problem, predicting employees' turnover across firms, is challenging and most of the existing approaches cannot adequately solve this problem. Using the Precision@30 results in Table 8 as an example, we observed that the Precision@30 of the baseline methods, GRU, TREE, LR, RF, SVM, DBGE, GBT, COX, AFT, and MTLR, is only up to 0.567 and HCPNN is 0.667; on the contrary, our method (i.e., CATCICE) yields 0.8, which indicates over 19.9% lift on the Precision@30. The results of other comparing groups are consistent with this observation. This observation implies that the problem of predicting employees' turnover across firms has not been effectively addressed by existing approaches and that a specifically designed method is indeed needed.

Second, our method outperforms the benchmark approaches across almost all groups of comparisons by a significant margin. For instance, as shown in Table 8, our method (i.e., CATCICE) yields 0.799 on NDCG@30, which is 15.5% higher than the best benchmark HCPNN (i.e., 0.692 on NDCG@30); our CATCICE outperforms HCPNN (the best benchmark) with a 25% improvement on Precision@50. Furthermore, the improvement in terms of Precision@k and NDCG@k is more significant compared to AUC, which indicates our method produces much more accurate turnover predictions in the top-k list of employees than the benchmark approaches do. Given that stakeholders like LinkedIn recruiter and company HR managers tend to focus on employees with the highest turnover likelihood, our CATCICE model would be more useful than the benchmark approaches in real-world practices. Notably, these results suggest that our proposed method is significantly more effective than the existing methods (especially GRU and HCPNN that employ RNNs to model sequential data) in modeling the temporal dynamics of the input factors and generating accurate predictions for employees' turnovers across firms. Thus, the results substantiate *Hypothesis 4* and confirm the superior efficacy of our CATCICE in comparison to existing methodologies.

| Methods | Prec@30 | Prec@50 | Prec@100 | NDCG@30 | NDCG@50 | NDCG@100 | AUC |
|---|---|---|---|---|---|---|---|
| **CATCICE**(Ours) | **0.800** | **0.700** | **0.640** | **0.799** | **0.727** | **0.672** | **0.704** |
| HCPNN | 0.667 | 0.560 | 0.550 | 0.692 | 0.607 | 0.580 | 0.691 |
| GRU | 0.467 | 0.520 | 0.500 | 0.493 | 0.523 | 0.507 | 0.684 |
| TREE | 0.567 | 0.560 | 0.450 | 0.581 | 0.534 | 0.487 | 0.672 |
| LR | 0.533 | 0.420 | 0.340 | 0.602 | 0.502 | 0.408 | 0.630 |
| RF | 0.533 | 0.480 | 0.480 | 0.526 | 0.486 | 0.486 | 0.692 |
| SVM | 0.100 | 0.060 | 0.060 | 0.092 | 0.065 | 0.063 | 0.538 |
| GBT | 0.533 | 0.500 | 0.430 | 0.505 | 0.487 | 0.438 | 0.696 |
| COX | 0.333 | 0.380 | 0.410 | 0.454 | 0.453 | 0.446 | 0.667 |
| AFT | 0.333 | 0.380 | 0.380 | 0.454 | 0.453 | 0.424 | 0.669 |
| MTLR | 0.067 | 0.080 | 0.080 | 0.100 | 0.100 | 0.092 | 0.533 |
| DBGE | 0.067 | 0.100 | 0.060 | 0.066 | 0.091 | 0.064 | 0.504 |

Table 8: Results with the $Test_6$ dataset (6-month prediction time period).

| Methods | Prec@30 | Prec@50 | Prec@100 | NDCG@30 | NDCG@50 | NDCG@100 | AUC |
|---|---|---|---|---|---|---|---|
| **CATCICE**(Ours) | **0.967** | **0.940** | **0.890** | **0.977** | **0.955** | **0.912** | **0.716** |
| HCPNN | 0.933 | 0.940 | 0.880 | 0.942 | 0.945 | 0.896 | 0.702 |
| GRU | 0.733 | 0.820 | 0.790 | 0.751 | 0.808 | 0.792 | 0.694 |
| TREE | 0.867 | 0.820 | 0.820 | 0.871 | 0.850 | 0.829 | 0.681 |
| LR | 0.800 | 0.620 | 0.510 | 0.831 | 0.695 | 0.581 | 0.640 |
| RF | 0.933 | 0.900 | 0.810 | 0.962 | 0.924 | 0.848 | 0.700 |
| SVM | 0.100 | 0.060 | 0.070 | 0.092 | 0.065 | 0.071 | 0.538 |
| GBT | 0.767 | 0.780 | 0.770 | 0.813 | 0.808 | 0.790 | 0.708 |
| COX | 0.833 | 0.840 | 0.790 | 0.865 | 0.860 | 0.813 | 0.677 |
| AFT | 0.833 | 0.820 | 0.780 | 0.861 | 0.842 | 0.802 | 0.679 |
| MTLR | 0.133 | 0.120 | 0.150 | 0.207 | 0.175 | 0.179 | 0.524 |
| DBGE | 0.100 | 0.120 | 0.090 | 0.089 | 0.108 | 0.090 | 0.502 |

Table 9: Results with the $Test_{12}$ dataset (12-month prediction time period).

### 5.3 Business Value of our CATCICE Model

In addition to the quantitative evaluations that demonstrate the superiority of our CATCICE model, this section presents a case study to illustrate its business value. Specifically, as online recruiters represent a critical stakeholder poised to benefit from turnover prediction, we demonstrate how the adoption of our CATCICE model could enable cost savings and improve operational efficiency for LinkedIn recruiters. Typically, online LinkedIn recruiters utilize InMail messages to contact candidates for available job positions (SocialTalent 2017), incurring a charge of $170 per month for up to 30 messages under a recruiter lite account, equating to $5.67 per InMail message[4]. Considering a scenario with 10 recruiters each contacting 30 candidates who have the highest turnover likelihood predicted by different methods, we estimate the number of responses that the recruiters could receive and the average cost per response. We assume that employees who are going to leave their current employers within six months will respond upon receiving InMail messages; and we infer the estimations based on the Precision@30 results of different methods on $Test_6$ (shown in Table 8).

The estimation results of the case study are presented in Table 10. Taking CATCICE in Table 10 as an example, the estimated total number of responses is 240 because (i) the Precision@30 of CATCICE is 0.8, and (ii) when each recruiter selects and contacts the top 30 candidates predicted by our CATCICE, 80% of the selected 300 candidates by all 10 recruiters will presumably respond. The average cost per response is estimated as $7.08 (i.e., $7.08=$1,700/240). On the contrary, HCPNN, the second-best competing method, yields 200 responses and $8.5 as the average cost per response. Compared to the benchmark methods, our CATCICE model yields substantially more responses, thus significantly improving recruiters' operational efficiency. As can be imagined, recruiters will save considerable amount of financial cost and time on searching for candidates by adopting our CATCICE model.

---

[4] https://www.linkedin.com/help/linkedin/answer/a417251/differences-between-recruiter-recruiter-professional-services-and-recruiter-lite

| Methods | Prec@30 | # of sent messages | # of responses | Total cost ($) | Cost per response ($) |
|---|---|---|---|---|---|
| **CATCICE** | **0.800** | **300** | **240** | **1700** | **7.08** |
| HCPNN | 0.667 | 300 | 200 | 1700 | 8.50 |
| GRU | 0.467 | 300 | 140 | 1700 | 12.14 |
| TREE | 0.567 | 300 | 170 | 1700 | 10 |
| LR | 0.533 | 300 | 160 | 1700 | 10.63 |
| RF | 0.533 | 300 | 160 | 1700 | 10.63 |
| SVM | 0.100 | 300 | 30 | 1700 | 56.67 |
| GBT | 0.533 | 300 | 160 | 1700 | 10.63 |
| COX | 0.333 | 300 | 100 | 1700 | 17 |
| AFT | 0.333 | 300 | 100 | 1700 | 17 |
| MTLR | 0.067 | 300 | 20 | 1700 | 85 |
| DBGE | 0.067 | 300 | 20 | 1700 | 85 |

Table 10: Results of the case study.

### 5.4 Importance of Different Components

To further investigate the importance of the employee factors (E), company factors (C), employee-company factors (EC), competitor influence (CI), and contagious effect (CE), we conducted an ablation study on the five components. This analysis evaluates the impact on prediction performance when one component is eliminated from our proposed method, CATCICE. Five variants of CATCICE were developed: CATCICE-E, CATCICE-C, CATCICE-EC, CATCICE-CI, and CATCICE-CE, each eliminating E, C, EC, CI and CE, respectively. The difference between our method with and without a removed component quantifies that component's contribution.

The prediction performance of these component-reduced models against CATCICE is presented in Table 11 for $Test_6$. The results clearly indicate the removal of each component leads to a decrease in prediction performance. Also, the removal of employee factors impairs CATCICE's performance the most among these five components. For instance, in Table 11, the removal of the employee factors reduces AUC, Precision@100, and NDCG@100 by 13.8%, 59.4%, and 57.7%, respectively. These ablation-study results suggest that all the five components make contributions to the prediction performance of our CATCICE model and that the employee factors are most important, thus supporting the *Hypotheses 1,2, and 3*. In Appendix A-7, we further show the results for $Test_{12}$,

| Methods | AUC | Prec@30 | Prec@50 | Prec@100 | NDCG@30 | NDCG@50 | NDCG@100 |
|---|---|---|---|---|---|---|---|
| CATCICE | 0.704 | 0.800 | 0.700 | 0.640 | 0.799 | 0.727 | 0.672 |
| CATCICE-E | 0.607 | 0.300 | 0.300 | 0.260 | 0.333 | 0.324 | 0.284 |
| CATCICE-C | 0.694 | 0.467 | 0.500 | 0.500 | 0.417 | 0.458 | 0.474 |
| CATCICE-EC | 0.697 | 0.600 | 0.600 | 0.560 | 0.606 | 0.605 | 0.573 |
| CATCICE-CI | 0.700 | 0.600 | 0.600 | 0.560 | 0.539 | 0.556 | 0.543 |
| CATCICE-CE | 0.700 | 0.633 | 0.660 | 0.620 | 0.707 | 0.705 | 0.657 |

Table 11: Results of ablation study with the $\text{Test}_6$ dataset.

## 5.5 Interpretation of Our Model

The business value of our CATCICE model lies in assisting stakeholders like HR managers and recruiters to identify employees with high turnover likelihood. However, they may hesitate to adopt such technical solutions due to their lack of interpretability, which complicates decision-making processes (Barredo Arrieta et al. 2020, Miller 2019). Interpreting how the input factors determine the predicted turnover likelihood can facilitate a better understanding among HR managers and recruiters about what drives an employee's turnover intention, thus encouraging them to adopt our solution and even tailor their retention or recruitment strategies accordingly. Gradient SHAP, a widely recognized measurement for attribution estimation, calculates SHAP values by approximating the expectations of model gradients (Lundberg and Lee 2017). We employed this technique to interpret the attributions of employee factors, company factors, employee-company factors, competitor influence, and contagious effect through using gradient SHAP for a positive (turnover) and a negative (staying) samples, respectively. The SHAP values are visualized in Figure 6 and Figure 7, where a higher SHAP value indicates a higher attribution of the input factors to the predicted score, and deeper red indicates a stronger turnover likelihood while deeper green indicates a stronger staying likelihood. Figure 6 interprets that the employee's turnover is strongly influenced by *Number of position changes in the past* in the 60th month. And the attributions of *Number of position changes in the past* in the 1st and 2nd months and *Average working months in previous companies* in the 60th month are slight. Conversely, *Number of job turnover in the past* in the 60th month negatively impact employee's turnover. In contrast, the staying intention of the negative sample, as illustrated in Figure 7, is positively correlated with *Number of job turnover in the past* in the 60th month, *Number of separations* in the 60th month, and *Number of months working in current company* in the 60th month.

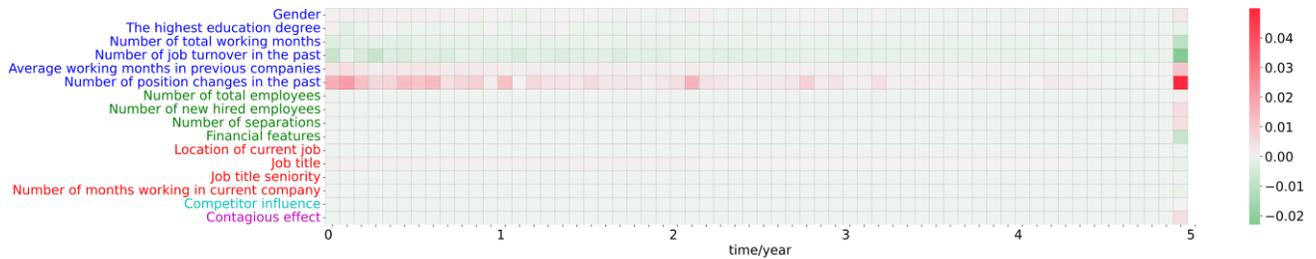

Figure 6: Feature attribution of a positive (turnover) employee.

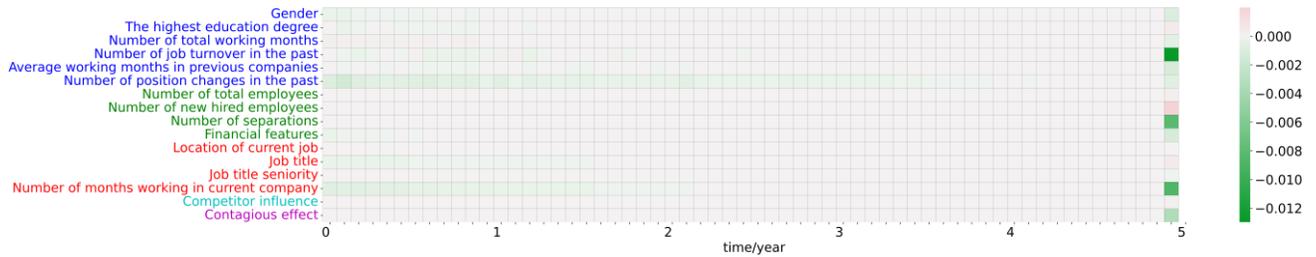

Figure 7: Feature attribution of a negative (survival) employee.

## 6. DISCUSSION AND CONCLUSION

In this study, we developed a novel transformer-based deep learning approach for predicting employees' turnover across different organizations by using public career data. Our method contrasts with existing methods that predict either turnover within a single firm or aggregated movement of employees among firms but fail to forecast individual employees' turnover across firms. Thus, our study fills the critical research gap in the literature on employee turnover prediction. This study makes several research contributions. *First*, we proposed a novel framework for predicting individual employee turnover across multiple firms. Rooted in theoretical foundations, this framework identifies and operationalizes three groups of driving factors and captures competitor influence and contagious effect to enhance prediction accuracy. *Second*, we developed a cross-component transformer model that consists of deeply customized cross-component attention and multi-head self-attention networks. Our method is capable of modeling time-dependencies within career trajectories and integrating contagious effect and competitor influence in employee turnovers, both of which are very effective for predicting employee turnover (as illustrated in our evaluation results). Our designed model is substantially different from existing transformer and cross-attention methods (as summarized in Section 3.3.4). *Third*, our research underscored the pivotal role of theoretical grounding in design science research within the domain of talent predictive analytics. Our design framework for employee turnover prediction is guided by the job embeddedness theory (Batt and Colvin 2011, Rubenstein et al. 2019, William

Lee et al. 2014). Thus, our study advocates for the application of theoretical frameworks in guiding the design of IT artifacts as a rigorous approach in design science research (Gregor and Hevner 2013, Hevner et al. 2004, Li et al. 2020). Through comprehensive experimental evaluations utilizing a real-world dataset, our method demonstrated superior performance over several state-of-the-art benchmark methods (e.g., 19.9% improvement of Precision@30 over the best benchmark method for 6-month turnover prediction). Moreover, our simulated cost saving for recruiters and interpretability of attributions of the input factors shed light on the practical insights of our turnover prediction solution.

Our study fits into several areas of IS research including turnover analysis (Ferratt et al. 2005, Joseph et al. 2012, 2015), machine learning and big data (Chen et al. 2012), and design science (Hevner et al. 2004, Rai 2017). *First*, turnover has attracted much attention in IS research. Most of the previous studies collect and utilize survey data to analyze factors such as job satisfaction and pay gap that drive employees' turnover (Joseph et al. 2007, 2012, 2015, Steel 2002). Recently, there has been a growing interest in leveraging massive talent data accumulated in various talent management IT systems such as LinkedIn and Careerbuilder to understand and predict employee movement (Li et al. 2017, Liu et al. 2020, Xu et al. 2019). These studies show great potential of using such talent data to develop decision support methods for talent management. Following these studies, we explored the publicly available employee profile data, operationalized a variety of factors, and proposed a novel cross-component attention model for predicting employee turnover. *Second*, this study contributes to the extant IS literature on big data and machine learning (Fang et al. 2013, Li et al. 2021, Padmanabhan et al. 2022). Developing better models and algorithms to tackle the analytical challenges of big data and address the business and societal problems is the focus of big data and machine learning research in IS literature (Chen et al. 2012, Padmanabhan et al. 2022). Consistent with the prior studies, we developed a novel machine learning method to tackle the prediction of employee turnover. Our developed model is able to capture the temporal dependencies and heterogeneity within massive career trajectory data and yield more accurate turnover prediction. *Third*, this study belongs to the computational genre of design science research (Rai 2017), the major objective of which is to develop novel data representations, computational algorithms, and analytics methods (Rai 2017). Extant computational design science studies have addressed various important domain problems that have business and societal impact (Fang et al. 2021). Following the design science research paradigm, we designed a kernel theory-driven computational approach for predicting employee turnover, which could benefit various business practices such as recruitment and employee retention.

Big data, machine learning and artificial intelligence have been recognized as crucial technologies for addressing many practical challenges in human resource management, talent analytics, and the future of work (Jain et al. 2018, NSF 2020). Employee turnover, a critical challenge for many organizations, incurs significant cost to individual organizations and broader economy. The big data and deep learning-driven solution for turnover prediction developed in this study has practical implications for various stakeholders. *First*, recruiters searching for potential hires may benefit from our solution by identifying individuals who most likely turn over soon. As we have demonstrated in the experimental evaluation (Section 5.3), our turnover prediction solution could significantly improve recruiters' operational efficiency and save them significant financial and time cost in talent search for vacant jobs. *Second*, online platforms such as LinkedIn and Indeed could benefit from accurate turnover predictions. A critical service of these online platforms is suggesting jobs and candidates to employees and employers (LinkedIn 2020, SocialTalent 2017). Current solutions mainly rely on the match between jobs and candidates. However, employees, who match the vacant positions well but have no desire to quit from their current jobs, will not apply for the recommended positions. If the predicted likelihood of turnover is considered in the current match-based solutions, it will improve the business performance of job and candidate recommendation services on the online platforms and further benefit workers and firms. *Third*, companies will benefit from knowing which employees will likely depart in advance. Average tenure of employees in most firms especially in IT industry is very low, and turnover has been a critical issue that causes huge cost (Mahan et al. 2019). With our turnover prediction solution, HR departments can proactively target employees likely to leave soon and take proactive actions to prevent their departure or to prepare for inevitable turnovers. Such proactive talent management, enabled by accurate turnover prediction, could significantly reduce the costs associated with employee turnover.

The current study can be extended in several directions. *First*, the inclusion of more features and variables could further improve the performance of the proposed transformer-based model. Beyond the public data used to operationalize the variables, various internal and private data collected by firms and organizations (such as employees' performance and evaluation records) contain valuable signals of turnover intentions which could be constructed and integrated into our developed method. *Second*, we trained and evaluated the learning model with the collected historical data. One direction worthy of future research is to study how to efficiently update the existing trained model with new data collected over time. As many turnovers across firms happen within one week or day, utilizing the new observations to extend the previously trained model's knowledge may continuously improve the prediction performance. *Third*, given that current job and candidate recommendation systems on many online career service platforms such as LinkedIn have not considered workers' turnover likelihood, it would be very interesting and practical to study how to integrate the turnover prediction model with recommendation methods such as deep collaborative filtering and then examine the potential improvement of recommendation performance.